\documentclass[10pt,twocolumn,letterpaper]{article}

\usepackage[pagenumbers]{cvpr}
\usepackage{algorithm}
\usepackage{algpseudocode}
\usepackage{multirow}
\usepackage{makecell}
\usepackage{xcolor}
\usepackage{booktabs}
\usepackage{colortbl}
\usepackage{graphicx}
\usepackage{array}
\usepackage{colortbl,xcolor}%

\usepackage[most]{tcolorbox}
\usepackage{tabularx} 
\usepackage{multirow} 
\usepackage{booktabs}
\usepackage{ragged2e}
\newcommand{\grayrow}{\rowcolor{gray!15}}
\definecolor{mybackblue}{RGB}{240,248,255}
\definecolor{myframeblue}{RGB}{100,149,237}
\newtcolorbox{keyfindingbox}[1][]{%
  enhanced,
  breakable,
  colback=mybackblue,
  colframe=myframeblue, 
  coltitle=white,
  fonttitle=\bfseries\small,
  attach boxed title to top left={yshift=-2mm, xshift=2mm},
  boxed title style={
    colback=myframeblue,
    sharp corners=south,
    boxrule=0pt, 
    arc=3pt,
    left=6pt,
    right=6pt,
    top=2pt,
    bottom=2pt,
  },
  title=#1,
  boxrule=0.4pt,
  arc=2mm,
  left=8pt,
  right=8pt,
  top=8pt,
  bottom=8pt 
}
\usepackage{etoolbox}
\usepackage{titletoc} 
\definecolor{cvprblue}{rgb}{0.21,0.49,0.74}
\usepackage[pagebackref,breaklinks,colorlinks,allcolors=cvprblue]{hyperref}

\def\paperID{}
\def\confName{CVPR}
\def\confYear{2026}

\title{Towards Reason-Informed Video Editing in Unified Models\\ with Self-Reflective Learning}
\author{%
\hspace{-.7em}%
Xinyu Liu$^{1,6}$,\;
\mbox{Hangjie Yuan$^{2}$\thanks{Project leader.},}\;
Yujie Wei$^{3,6}$,\;
Jiazheng Xing$^{2}$,\;
Yujin Han$^{4,6}$,\;
Jiahao Pan$^{1}$,\;
Yanbiao Ma$^{5}$,\;\\[0.1em]
\hspace{-.7em}Chi-Min Chan$^{1}$,\;
Kang Zhao$^{6}$,\;
Shiwei Zhang$^{6}$,\;
Wenhan Luo$^{1}$\thanks{Corresponding author.},\;
Yike Guo$^{1}$ \\[0.2em]
{$^1$HKUST}\quad
{$^2$ZJU}\quad{$^3$FDU}\quad{$^4$HKU}\quad{$^5$RUC}\quad{$^6$Tongyi Lab}
}
\begin{document}
\maketitle
\begin{abstract}
Unified video models exhibit strong capabilities in understanding and generation, yet they struggle with reason-informed visual editing even when equipped with powerful internal vision-language models (VLMs). We attribute this gap to two factors: (1) existing datasets are inadequate for training and evaluating reasoning-aware video editing, and (2) an inherent disconnect between the models’ reasoning and editing capabilities, which prevents understanding from guiding the editing process.
To address this, we introduce the Reason-Informed Video Editing (RVE) task, which requires reasoning about physical plausibility and causal dynamics during editing. To support systematic evaluation, we construct RVE-Bench, a comprehensive benchmark with two complementary subsets: Reasoning-Aware Video Editing (RAVE) and In-Context Video-to-Video Generation (ICVG), spanning diverse reasoning dimensions across both editing and generation scenarios.
Building upon this foundation, we propose ReViSE, a self-reflective learning framework that harnesses the model's internal VLM to evaluate and refine its own generation during training. Unlike prior reward-based approaches that rely on external critics, ReViSE leverages the model’s internal VLM as a self-reflective evaluator, providing differentiable feedback that directly refines the generator’s reasoning behavior during training.
Extensive experiments on RVE-Bench demonstrate that ReViSE enhances editing accuracy and visual fidelity, outperforming the finetuned counterpart by \textbf{10\%} in Overall score on the RAVE subset, demonstrating the effectiveness of self-reflective differentiable reward.
\end{abstract}    
\section{Introduction}
\label{sec:intro}
\begin{figure*}[t]
  \centering
\includegraphics[width=1.0\linewidth]{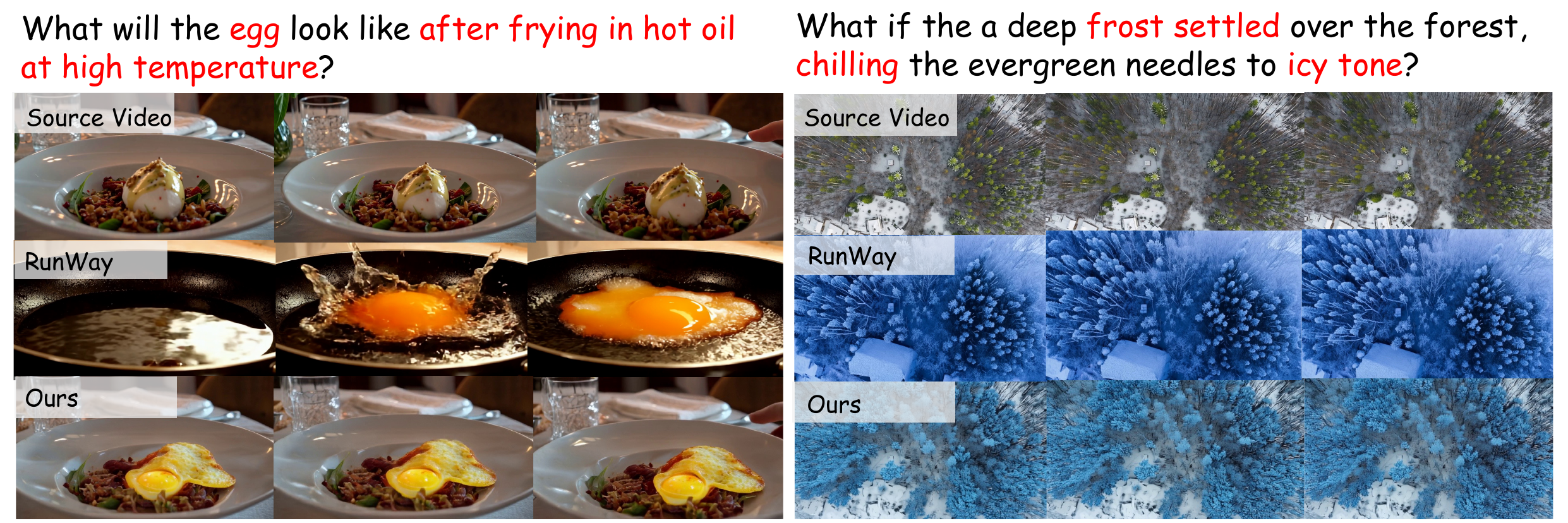}
  \caption{
    {Comparison between ReViSE and Runway Gen-4 Aleph~\cite{runway2025aleph}. 
Runway fails to generate a fried egg and loses background consistency with the source 
(left), and disrupts the scene layout by over-editing unintended regions (right). 
ReViSE correctly reasons about the intended edit in both cases.}}
  \label{fig:runway}
\end{figure*}

\begin{figure*}[t]
  \centering
   \includegraphics[width=1\linewidth]{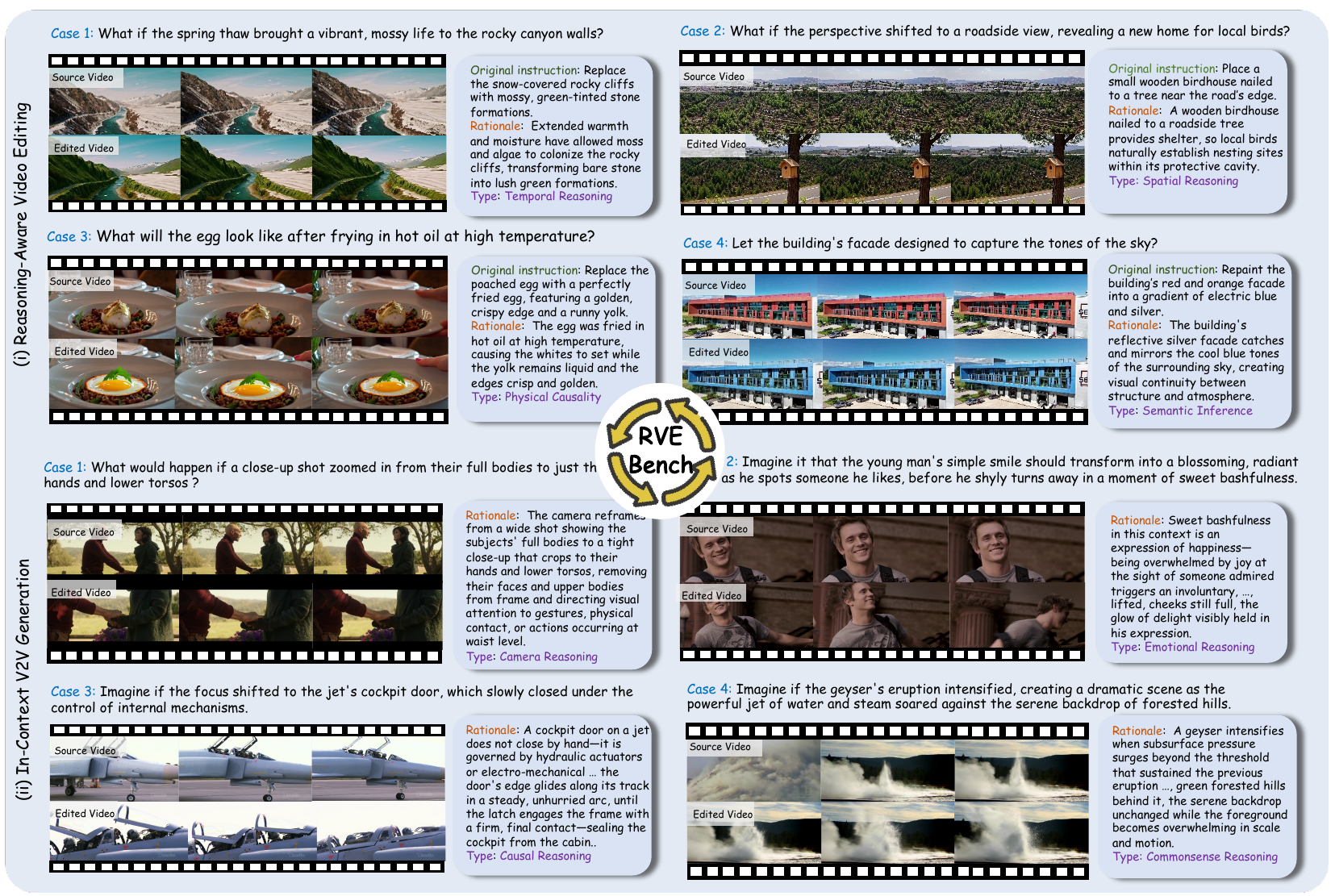}
   \caption{
    {\textbf{Overview of RVE-Bench.} Representative examples from the two subsets: RAVE (top) and ICVG (bottom), each covering four reasoning types.}}
   \label{fig:rve_bench}
\end{figure*}

Instruction-guided video editing~\cite{bai2025scaling,qin2024instructvid2vid,zhang2024effived,chai2023stablevideo,wu2024fairy,ceylan2023pix2video,xing2023vidiff} has advanced rapidly with unified generation models~\cite{tan2025omni,pan2025transfer,dong2023dreamllm}, yet existing approaches remain confined to \emph{literal} editing, where the target visual state is fully specified in the text (\eg, \texttt{remove the boat}).
Real-world editing, however, frequently demands \emph{reason-informed} transformations where the intended outcome is underspecified and must be inferred from physical laws, temporal dynamics, or world knowledge (\eg, \texttt{Imagine the scene an hour after the boat departed}).
We term this setting \textbf{Reason-Informed Video Editing (RVE)}, spanning physical, causal, temporal, and narrative reasoning dimensions, a capability axis that remains largely unexplored.
As Figure~\ref{fig:runway} illustrates, this gap persists even at the commercial frontier: Runway Gen-4 Aleph~\cite{runway2025aleph}, one of the most capable commercial video generation systems available, fails to correctly execute RVE instructions, over-editing by modifying unintended subjects and disrupting scene layout.
This demonstrates that reason-informed video editing represents a deep and underexplored challenge unresolved by current state-of-the-art systems.
To enable systematic study of this gap, we introduce \textbf{RVE-Bench}, the first benchmark dedicated to reason-informed video editing, comprising approximately 1{,}000 evaluation triplets across two complementary subsets~(Figure~\ref{fig:rve_bench}).
\textbf{RAVE} (Reasoning-Aware Video Editing) requires the model to apply targeted modifications \texttt{within} a source video, where the correct edit is not visually explicit but must be inferred through physical, causal, or world-knowledge reasoning.
\textbf{ICVG} (In-Context V2V Generation) requires the model to generate a plausible \texttt{new} segment given a source video clip, inferring the implicit narrative or causal intent to produce coherent future frames rather than editing existing ones.
To support model training, we further construct \textbf{RVE-Dataset}, a large-scale corpus of 56K reasoning-enriched triplets built via a scalable, fully automated pipeline.

Training on the RVE-Dataset substantially improves RVE performance, yet the model remains bounded by output-level reconstruction loss, which provides no instruction-specific signal about whether an edit correctly satisfies the underlying reasoning requirement.
Since unified video models already embed a VLM capable of multimodal reasoning, a natural extension is to exploit this comprehension capacity directly, letting the model critique its own outputs in natural language to form a self-reflective training signal.
However, such text-based feedback yields discrete language-level signals that are non-differentiable, offering no gradient path to the generator.
We address this with \textbf{ReViSE}, a self-reflective learning framework that empowers unified video models to refine their reasoning through intrinsic differentiable feedback.
Formulating edit quality as a QA task and using the VLM's token-level ``Yes'' probability as the reward is architecturally homogeneous with the VLM's native token-prediction objective, yielding a continuous, differentiable signal without any additional regression head.
On held-out pairs, $P(\text{``Yes''})$ achieves a higher Spearman correlation with editing accuracy~(Figure~\ref{fig:scatter}), substantially outperforming hard-decoded binary signals and confirming the frozen VLM as an effective reward proxy.
The critic evaluates each edit across four dimensions spanning edit fidelity, content preservation, temporal smoothness, and visual realism, preceded by chain-of-thought reasoning to ground the verdict in interpretable evidence.
Crucially, since the VLM is already embedded in the unified architecture, this reward is entirely intrinsic, requiring neither labeled preference data nor any additional inference cost at deployment.
Gradients flow directly from the frozen VLM's token logits through a one-step clean video estimate to the generator, forming a fully differentiable self-corrective loop, unlike RL-based alignment methods~\cite{black2023training,liu2025flow,xue2025dancegrpo} that depend on costly external reward models or policy-gradient estimators.
Jointly optimized with flow-matching, ReViSE achieves state-of-the-art performance across all reasoning categories on RVE-Bench.
Our contributions are summarized as follows:
\begin{itemize}
    \item We introduce \textbf{RVE-Bench}, the \textbf{first} benchmark dedicated to reason-informed video editing spanning RAVE and ICVG subsets, along with a scalable automated pipeline that constructs \textbf{RVE-Dataset}, a large-scale collection of 56K reason-informed training triplets.
    \item We propose \textbf{ReViSE}, a self-reflective learning framework that enhances unified video models' reason-informed editing capability via a differentiable intrinsic reward derived from the model's own VLM, enabling reason-informed training without external reward models, labeled preference data, or additional inference cost.
    \item We conduct extensive experiments demonstrating state-of-the-art performance on RVE-Bench across all reasoning categories and sustained superiority on cross-dataset generalization and literal editing benchmarks.
\end{itemize}
\section{Related Work}
\subsection{Instructional Video Editing}
Recent advances in video editing~\cite{yang2025videograin, ceylan2023pix2video, liu2024video, chai2023stablevideo,ku2024anyv2v,ouyang2024i2vedit,bian2025videopainter,qi2023fatezero,sun2024generative,tu2025videoanydoor,wang2024lave,ye2025unic,zhuang2025get,zi2025cococo} have leveraged diffusion-based generative models~\cite{yan2021videogpt, hong2023cogvideo, esser2023structure, mei2023vidm, chai2023stablevideo, ceylan2023pix2video, blattmann2023svd, wu2023tune, brooks2024video,wu2025omnigen2,wei2024dreamvideo,wei2025dreamrelation,wei2024dreamvideo2,han2025can,chen2025masked} to achieve high-quality, temporally consistent, and controllable video synthesis. A growing body of research focuses on extending image diffusion models to the video domain, emphasizing instruction following and adaptation efficiency. Early efforts such as Tune-A-Video~\cite{wu2023tune} pioneered one-shot tuning of pre-trained image diffusion models for text-to-video generation. To enhance temporal alignment and domain generalization, InsV2V~\cite{cheng2023consistent} introduces a video-to-video transfer framework using synthetic datasets for improved stability. Recently, large-scale synthesized datasets such as InsViE~\cite{wu2025insvie}, Senorita\text{-}2M~\cite{zi2025se} and Ditto~\cite{bai2025scaling} have been introduced to facilitate instruction-guided video editing, providing diverse and high-quality examples for training and evaluation.
Beyond isolated models, unified frameworks have been explored to integrate VLMs to enhance instruction comprehension and cross-modal grounding for editing. For instance, VEGGIE~\cite{yu2025veggie} parses complex natural-language instructions and reasons over spatio-temporal regions to produce semantically consistent, context-aware edits; meanwhile, Omni-Video~\cite{tan2025omni} couples VLM-based instruction interpretation with diffusion-based generation in a single end-to-end trainable system, improving faithfulness in instruction following and alignment of edits to semantic intent.
\subsection{Reason-Informed Visual Generation and Editing}
Recent efforts in reason-informed visual generation and editing~\cite{wiedemer2025video,niu2025wise,chen2025r2i,zhang2025worldgenbench,yang2024editworld,li2025science,cong2025viva} focus on bridging the gap between perceptual quality and semantic coherence. Benchmarks like WISE~\cite{niu2025wise}, R2I-Bench~\cite{chen2025r2i}, and WorldGenBench~\cite {zhang2025worldgenbench} evaluate text-to-image models' ability to reason about world knowledge, spatial relationships, and causal logic, revealing widespread failures in factual accuracy and commonsense understanding. RISE~\cite{zhao2025envisioning} and EditWorld~\cite{yang2024editworld} extend this to image editing, assessing whether edits preserve physical plausibility and contextual consistency, while Science-t2i~\cite{li2025science} specifically targets scientific accuracy in generated images. Together, these works advance the frontier of reasoning-grounded visual generation.
Recent Reinforcement Learning methods~\cite{han2025turning,zhou2024calibrated,jin2025srum,choi2024self,fu2020sscr,luo2025dual,kumari2025learning} have focused on improving model alignment without relying on external reward models. Methods like CSR~\cite{zhou2024calibrated} leverage internal consistency and cross-modal alignment as implicit rewards, enabling iterative self-improvement in multimodal understanding. However, these approaches target general alignment rather than reason-informed visual editing.

More closely related are differentiable reward learning methods~\cite{xu2023imagereward,prabhudesai2024video,gong2025onereward,wu2025rewarddance,ding2025dollar} that backpropagate reward signals directly through the generator during training. ReFL~\cite{xu2023imagereward} and VADER~\cite{prabhudesai2024video} optimize image and video diffusion models via pretrained external reward models. RewardDance~\cite{wu2025rewarddance} reframes reward as the VLM's ``Yes'' token probability, aligning reward with next-token prediction. However, all these methods rely on external reward models that are costly to obtain and may not capture reasoning-level semantics. ReViSE instead uses the model's own frozen VLM as an intrinsic differentiable reward, requiring no external critic, labeled preference data, or additional inference cost.
\section{RVE-Dataset and RVE-Bench}
\subsection{Data Curation}
We introduce the RVE-Dataset, a large-scale collection of reasoning-enriched samples designed to foster sophisticated video editing reasoning. The dataset features two subsets, each providing source/target video pairs coupled with reasoning-aware instructions and detailed type annotations. Specifically, the RAVE subset spans four reasoning types (Temporal, Spatial, Physical Causality, and Semantic Inference), while the ICVG subset spans four complementary types (Camera, Causal, Emotional, and Commonsense Reasoning).
\subsubsection{Reasoning-Aware Video Editing (RAVE)}
This subset is built upon existing instruction-based video editing datasets (e.g., Ditto-1M~\cite{bai2025scaling}).
The original instructions describe surface-level pixel manipulations but lack the implicit reasoning required for real-world understanding.
We employ GPT-4o~\cite{hurst2024gpt} to rewrite each instruction by inferring a real-world rationale and reformulating it as an implicit reasoning cue that withholds the literal operation while remaining consistent with the source and target videos.
To ensure instruction quality, we first manually annotate a small set of examples covering all reasoning types, which serve as few-shot references to guide VLM toward physically motivated and semantically coherent formulations.
As illustrated in Figure~\ref{fig:rve_bench}, Case 1 (Temporal Reasoning) grounds a snow-to-moss edit in the ecological process of spring thaw enabling moss colonization, while Case 4 (Semantic Inference) reframes a color change as the natural reflective property of a silver facade mirroring the surrounding sky, demonstrating that the rewriting is physically motivated and logically consistent with the visual change.
This yields high-quality reasoning-aware triplets spanning four types: Temporal Reasoning, Spatial Reasoning, Physical Causality, and Semantic Inference.
\subsubsection{In-Context V2V Generation (ICVG)}
\begin{figure}[t]
  \centering
\includegraphics[width=\linewidth]{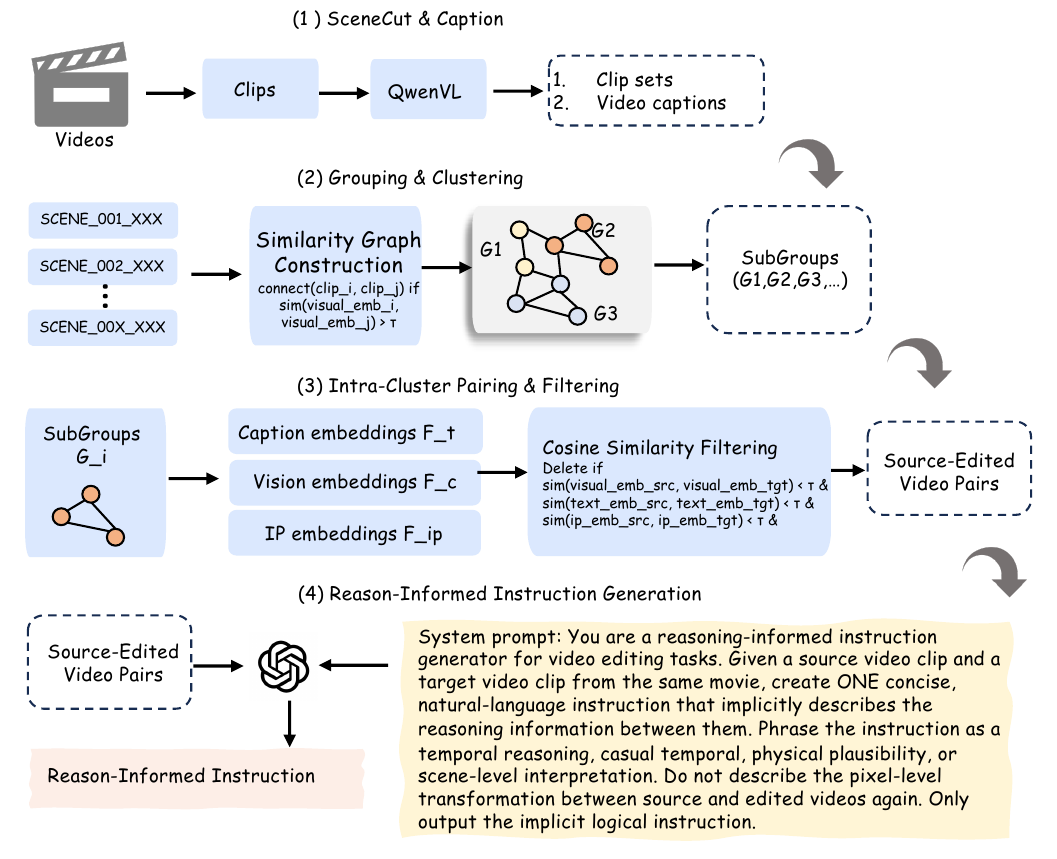}
  \caption{Overview of the data construction pipeline for the ICVG subset.}
  \label{fig:dataset_pipeline}
\end{figure}

While the first subset targets instruction-guided editing of existing content, the second focuses on generating plausible future video frames conditioned on contextual descriptions.
As illustrated in Figure~\ref{fig:dataset_pipeline}, we construct this subset directly from movie data, leveraging their naturally rich temporal continuity and semantic diversity.
Each movie is segmented into shots~\cite{yu2001efficient} and captioned with QwenVL-32B~\cite{bai2025qwen2}.
Clips from the same movie 
are grouped into consecutive scene windows, 
within which visual clustering is performed using SigLIP~\cite{tschannen2025siglip} embeddings, where clips whose pairwise cosine similarity exceeds a data-driven threshold are connected to form visually coherent sub-clusters.
Within each sub-cluster, source and target pairs are selected by a scoring function that jointly maximizes semantic similarity between captions and visual diversity between clips, with the earlier clip in temporal order serving as source and the later as target.
A visual consistency filter is applied to ensure each pair shares sufficient contextual overlap.
For each selected triplet, we utilize GPT-4o to generate reason-informed instructions grounded in narrative, causal, or commonsense reasoning, covering four types including Camera, Causal, Emotional, and Commonsense Reasoning.
Unlike RAVE, GPT-4o synthesizes the instruction from scratch by inferring a plausible event or perspective shift that connects the two clips.
As illustrated in Figure~\ref{fig:rve_bench} (Camera Reasoning), the synthesized instructions are narratively motivated and semantically coherent with the target clip.
By combining the rewritten reasoning-enriched instructions of the RAVE subset with these synthesized instructions over real movie clips, our dataset achieves both visual diversity and semantic depth, providing a strong foundation for training models capable of reason-aware and physically plausible video editing.
Across both subsets, GPT-4o is used exclusively for text instruction generation and never influences the visual content, as the source/target video pairs in RAVE originate from human-annotated data and those in ICVG are selected by an algorithmic pipeline based on visual clustering, effectively decoupling data construction from automated evaluation.
\subsection{Benchmark Construction}
\label{sec:rve_bench_eval_metrics}
To facilitate systematic evaluation of the RVE task, we construct RVE-Bench, the first comprehensive benchmark designed for this task.
RVE-Bench is built by holding out a dedicated test split from the RAVE and ICVG data sources, ensuring strict separation between training and evaluation samples to prevent data leakage.
It comprises approximately 1,000 unique triplets, each containing a source video, a textual instruction, and a corresponding edited video. All benchmark instructions are manually verified to ensure physical plausibility and alignment with the visual content.
Inspired by recent work in automated evaluation~\cite{ku2024viescore}, we propose a robust, reasoning-aware evaluation framework utilizing GPT-4o.
Our framework assesses generated videos along two aspects, \textit{Semantic Consistency} (SC) and \textit{Perceptual Quality} (PQ), where SC measures how faithfully the edit adheres to the instruction and PQ evaluates the visual and temporal integrity of the edited video.
Each aspect is further broken down into two sub-metrics.
\begin{enumerate}
    \item \textbf{Edit Accuracy (EA)}: evaluates whether the edit aligns with the instructions.
    \item \textbf{Preservation Consistency (PC)}: assesses the consistency of non-edited regions with the source video to avoid over-editing.
    \item \textbf{Generation Naturalness (GN)}: evaluates the smoothness and natural flow of the video.
    \item \textbf{Generation Realism (GR)}: measures the visual fidelity of the generated frames, focusing on the absence of artifacts, distortions, or unnatural textures.
\end{enumerate}
For each pair, GPT-4o provides a score from 0 to 10 for each applicable sub-metric. The Overall score for a method is computed by first taking the geometric mean of the SC and PQ sub-metrics, and then averaging across the benchmark. For the ICVG task, the PC metric is not applicable, as ICVG involves generating new content conditioned on context rather than applying targeted edits, making the boundary between edited and non-edited regions undefined.
\section{Method}
\begin{figure}[t]
  \centering
  \includegraphics[width=\linewidth]{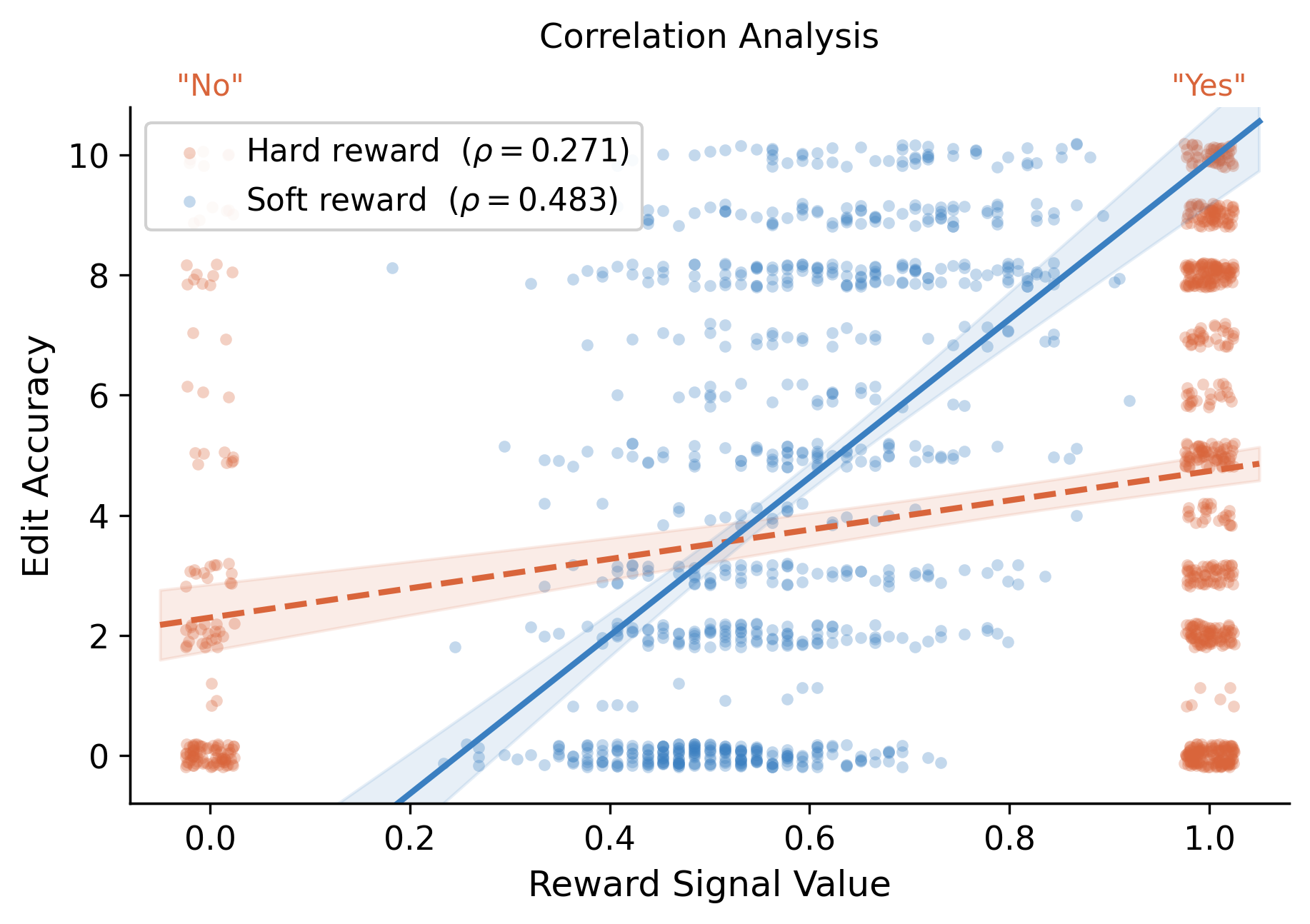}
  \caption{Correlation between reward signals and edit accuracy.}
  \label{fig:scatter}
\end{figure}
This section presents \textbf{ReViSE}, a self-reflective framework that bridges the reasoning gap between VLM understanding and generator editing in unified video models. We first introduce the base architecture (Sec.~\ref{sec:preliminary}), then detail the critic mechanism and the optimization objective (Sec.~\ref{sec:self_reflective_learning}).
\subsection{Preliminary}
\label{sec:preliminary}
Connector-based unified models~\cite{tan2025omni} couple a VLM 
$\mathcal{U}$ (e.g., ViLA~\cite{wu2024vila}) with a diffusion 
generator $G_\phi$ (e.g., Wan~\cite{wan2025wan}) through 
connector $f_\mathcal{C}$. Given source video $y_i$ and instruction $c_i$, the video encoder and T5 
encoder extract visual and textual representations, while $\mathcal{U}$ 
produces a multimodal semantic representation:
\begin{equation}
\mathbf{v}_i = E_{\text{vid}}(y_i),\quad 
\mathbf{t}_i = E_{\text{text}}(c_i),\quad 
\mathbf{u}_i = \mathcal{U}(y_i,\,c_i).
\end{equation}
These are fused by $f_{\mathcal{C}}$ into a unified conditioning signal 
$\mathbf{c}_i = f_{\mathcal{C}}(\mathbf{v}_i, \mathbf{t}_i, \mathbf{u}_i)$, 
which is injected into each DiT block~\cite{peebles2023scalable} to guide 
generation. 
Supervised finetuning follows flow matching:
\begin{equation}
\mathcal{L}_{\text{FM}} = \mathbb{E}_{t,\epsilon}\bigl[\|\mathbf{v}_\phi(z_t, t \mid \mathbf{c}_i) - (\epsilon - x_0)\|^2\bigr],
\end{equation}
with $z_t = (1-t)x_0 + t\epsilon$. 
We observe that baseline VLMs exhibit stronger reasoning capabilities for evaluating reason-informed edits than their generators do for producing them. However, connector-based architectures create a fundamental \emph{reasoning-generation disconnect}: the connector transmits VLM representations but not reasoning constraints to $\mathcal{L}_{\text{FM}}$. Standard SFT exacerbates this disconnect by optimizing only output similarity, providing no signals about instruction reasoning requirements. Consequently, models learn to \emph{imitate} visually plausible edits without physical or causal reasoning capability.
\begin{figure*}[t]
  \centering
   \includegraphics[width=1.0\linewidth]{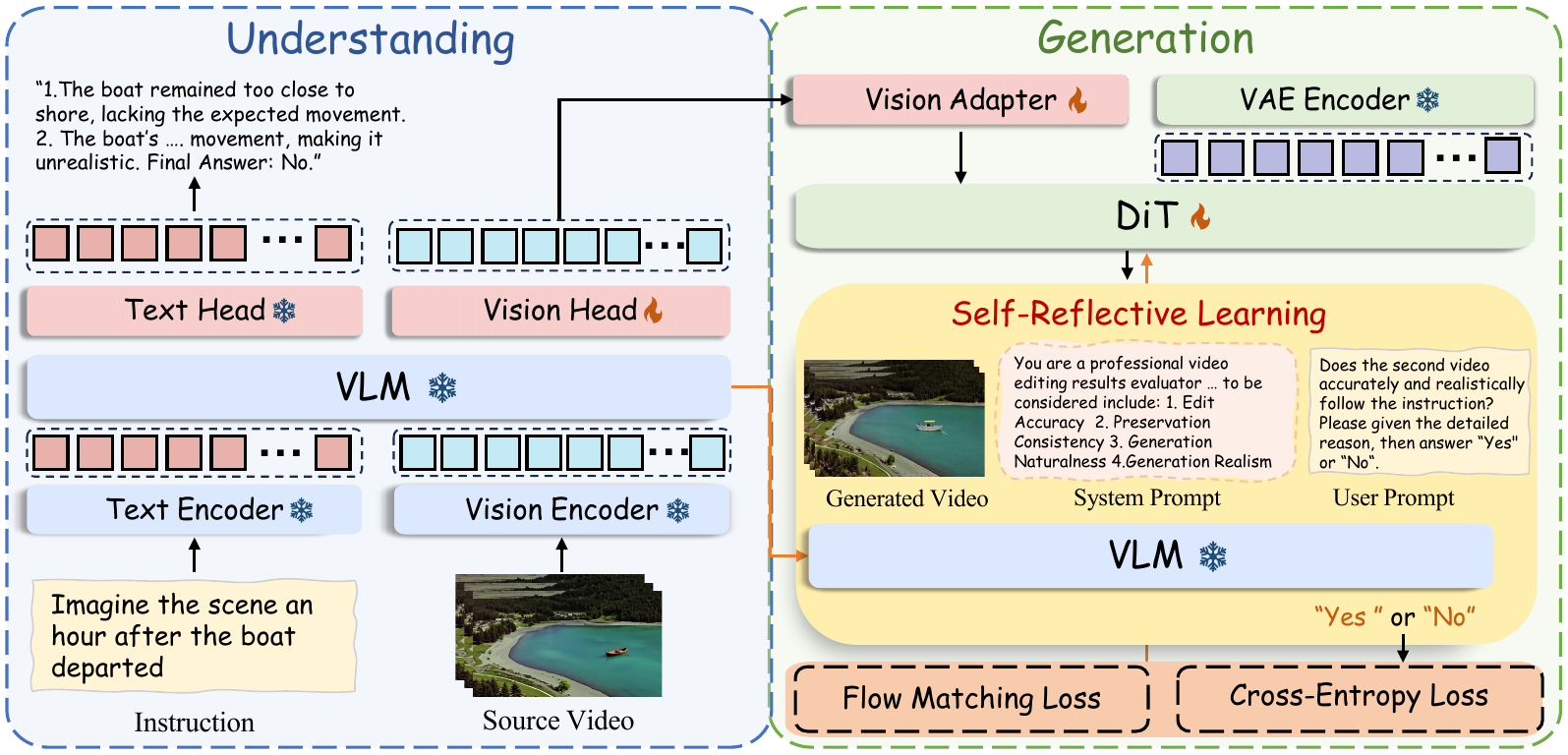}
   \caption{
   {
   \textbf{Overview of ReViSE's framework.} Given a reasoning instruction and source video, the internal VLM produces reasoning tokens and visual features that condition a DiT generator via a vision head and adapter. The frozen VLM then evaluates the generated output across four quality dimensions by answering ``Yes'' or ``No'', producing token-level reward $P(\text{``Yes''})$ that backpropagates directly into the generator, closing the reasoning-generation gap without any external critic.
   }
   }
   \label{fig:main}
\end{figure*}
\subsection{Self-Reflective Learning}
\label{sec:self_reflective_learning}
To bridge the reasoning-generation disconnect, ReViSE introduces a
self-reflective training loop in which $G_\phi$ generates an edited video,
frozen $\mathcal{U}$ evaluates whether the edit satisfies the reasoning
instruction, and the resulting signal is backpropagated end-to-end to
$G_\phi$. Crucially, the gradient flows directly from the VLM token logits through the frozen VAE decoder $\mathcal{D}$ and the clean estimate $\hat{x}_0$ back to the generator parameters $\phi$, forming a globally differentiable reward path that requires neither policy-gradient estimators nor external reward models.
\subsubsection{Clean Video Estimation.}
A key challenge is that training operates on noisy latents $z_t$, which 
are semantically uninterpretable for $\mathcal{U}$; we therefore first 
derive a one-step clean estimate before critic evaluation, following the 
same insight as prior work~\cite{luo2025dual}.
Formally, given $z_t = (1-t)x_0 + t\epsilon$ and predicted velocity 
$\mathbf{v}_\phi \approx \epsilon - x_0$, we recover:
\begin{equation}
  \hat{x}_0 = z_t - t\cdot\mathbf{v}_{\phi}(z_t, t \mid \mathbf{c}_i),
\end{equation}
which provides $\mathcal{U}$ with a semantically coherent input without 
requiring a multi-step denoising rollout.
However, at high noise levels ($t$ close to 1) the one-step estimate
carries insufficient visual semantics for reliable critic evaluation.
We therefore gate the VLM loss to timesteps where $t \leq \sigma_{\max}$,
falling back to standard flow matching otherwise, and set
$\sigma_{\max} = 0.5$ empirically.
We uniformly sample $F$ frames from $\hat{x}_0$ and decode them via the frozen VAE decoder as $\hat{y}_0 = \mathcal{D}(\hat{x}_0^{(F)}) \in \mathbb{R}^{F \times 352 \times 640 \times 3}$, reducing VAE decoding overhead while retaining sufficient temporal content for critic evaluation.
\subsubsection{Self-Reflective Critic.}
Given the clean estimate $\hat{y}_0$, we require a differentiable signal that indicates whether the proposed edit satisfies the reasoning constraints of the instruction. A natural baseline is to prompt $\mathcal{U}$ for a ``Yes''/``No'' 
verdict and use the decoded token as a binary reward; however, 
decoding collapses the full probability distribution into 
a single discrete value, discarding the model's uncertainty and 
confidence gradients.
We instead use $\mathcal{U}$'s token-level probability $P(\text{``Yes''})$ 
directly as the reward signal: soft logits preserve the 
reasoning-level semantics that hard decoding discards.
We verify this empirically on 800 held-out pairs with human-annotated edit accuracy labels in Figure~\ref{fig:scatter}. $P(\text{``Yes''})$ achieves a Spearman correlation of $\rho{=}0.483$ ($p{<}10^{-47}$) with human edit accuracy, compared to $\rho{=}0.271$ ($p{<}10^{-14}$) for hard-decoded binary outputs, confirming that soft token probabilities retain reasoning-level semantics discarded by hard decoding.
We further validate the GPT-4o-based evaluation protocol on held-out RVE-Bench samples, finding a Spearman correlation of $\rho{=}0.761$ between GPT-4o scores and human expert ratings, confirming the reliability of our automated evaluation framework.

Concretely, given $(\hat{y}_0, y_i, c_i)$ and system prompt $S_q$ 
(Figure~\ref{fig:main}), $\mathcal{U}$ is directed to evaluate the edit 
across four key dimensions: \textbf{(1) edit accuracy}; 
\textbf{(2) preservation consistency}; 
\textbf{(3) generation naturalness}; 
\textbf{(4) generation realism}, as detailed in 
Sec.~\ref{sec:rve_bench_eval_metrics}.
The VLM responds ``Yes'' only if all four dimensions are satisfied; 
otherwise it responds ``No''.
To enhance interpretability, $\mathcal{U}$ is further instructed to 
first produce a concise \textit{chain-of-thought reasoning} before 
the final verdict. We extract the logit difference between ``Yes'' 
and ``No'' tokens as the differentiable reward signal.
\subsubsection{Training Objective.}
ReViSE enhances the standard training objective by incorporating an auxiliary reasoning loss from the critic. The total loss combines the flow-matching loss \(\mathcal{L}_{\text{FM}}\) with the proposed reasoning loss \(\mathcal{L}_{\text{reason}}\).
The reasoning loss \(\mathcal{L}_{\text{reason}}\) is defined as a binary cross-entropy loss based on the logit difference between the ``Yes'' and ``No'' tokens:
\begin{equation}
\mathcal{L}_{\text{reason}} = - \log p(a_j), \quad \text{where} \quad p(a_j) = \sigma(\ell^{(j)}_{a_j} - \ell^{(j)}_{\bar{a}_j}),
\end{equation}
where \( \ell^{(j)}_{a_j} \) and \( \ell^{(j)}_{\bar{a}_j} \) denote the logits of the correct and opposite answer tokens, respectively, and \( \sigma(\cdot) \) is the sigmoid function. This formulation normalizes the logits within the ``yes'' and ``no'' space, ensuring stability and effectiveness during training.
Since a pretrained VLM may assign non-trivial yes probability even to an unedited source video, the absolute logit difference conflates this unconditional bias with genuine editing quality.
To isolate the editing contribution, we subtract a no-gradient baseline logit difference obtained by feeding the self-comparison pair $(y_i, y_i)$ through $\mathcal{U}$.
The resulting \textbf{relative} logit difference $\ell^{(j)}_{a_j} - \ell^{(j)}_{\bar{a}_j}$ then measures how much more convincingly the generated $\hat{y}_0$ satisfies the instruction \emph{compared to leaving the source unchanged}, and is positive only when the edit produces a genuine improvement, yielding a zero-centered, bias-free reward signal.
Importantly, the parameters of both \(\mathcal{U}\) and \(\mathcal{D}\) are kept \textbf{frozen} throughout training; only the generator parameters \(\phi\) are updated.
Gradients from \(\mathcal{L}_{\text{reason}}\) propagate through the frozen VLM forward pass, the frozen VAE decoder \(\mathcal{D}\), and the clean estimate \(\hat{x}_0\) back to \(\phi\), so the generator learns to produce outputs that its internal critic would evaluate as correct.
The total loss, which we term Unified Semantic Optimization (USO), is:
\begin{equation}
\mathcal{L}_{\text{USO}} = \mathcal{L}_{\text{FM}} + \lambda \cdot \mathcal{L}_{\text{reason}},
\end{equation}
where $\lambda$ balances generative fidelity against reasoning-level alignment.
This USO objective jointly optimizes for generative fidelity via \(\mathcal{L}_{\text{FM}}\) and high-level semantic alignment via \(\mathcal{L}_{\text{reason}}\).
Since \(\mathcal{U}\) is frozen, its evaluation criteria remain stable and cannot be gamed by the generator, which avoids the reward hacking problem that would arise if both modules were jointly optimized.
\section{Experiments}
\subsection{Implementation Details} 
We build ReViSE upon the Omni-Video~\cite{tan2025omni} framework, which couples a frozen VLM ($\mathcal{U}$, 11B parameters) with a Wan2.1-1.3B DiT as the generator $G_\phi$.
We fine-tune $G_\phi$ and the visual context adapter using AdamW with a learning rate of $3\times10^{-6}$, weight decay of $10^{-4}$, gradient clipping of $0.1$, and a batch size of 8.
We set the reasoning loss weight $\lambda=0.25$, the noise-gating threshold $\sigma_{\max}=0.5$, and the classifier-free guidance drop ratio to $0.2$. While VLM scoring introduces training overhead compared to SFT, activating the self-reflective step only at low-noise timesteps limits the amortized per-iteration latency to $5.09\,\text{s/it}$, substantially lower than policy-gradient methods~\cite{black2023training,liu2025flow,xue2025dancegrpo} that require full denoising rollouts and reward queries at every step. ReViSE preserves Omni-Video's inference FLOPs ($3.513\,\text{P/step}$), adding zero deployment overhead.

\subsection{Baselines and Evaluation} 
To evaluate our method's reasoning-informed video editing capabilities, we utilize the RVE-Bench benchmark comprising $\sim$1K test samples across diverse reasoning categories. We compare ReViSE with four state-of-the-art baselines, including two diffusion-based instruction editing methods, InsV2V~\cite{cheng2023consistent} and InsViE~\cite{wu2025insvie}, one controllable editing pipeline, VACE~\cite{jiang2025vace}, and one unified video model, Omni-Video~\cite{tan2025omni}. The results are evaluated across two complementary metric groups as reported in Table~\ref{tab:main_result}. The first group covers VLM evaluation scores comprising Edit Accuracy (EA), Preservation Consistency (PC), Generation Naturalness (GN), Generation Realism (GR), and Overall score. The second group covers video quality metrics, including ViCLIP\textsubscript{T}~\cite{wang2023internvid} and five VBench~\cite{huang2024vbench} metrics, namely Subject Consistency (SC), Background Consistency (BC), Overall Consistency (OC), Temporal Flickering (TF), and Motion Smoothness (MS), providing a comprehensive assessment of both semantic editing performance and perceptual video quality. Note that PC is not reported for the ICVG subset, as its instructions often require substantial transformations where strict preservation of the source is neither expected nor meaningful.
\subsection{Experimental Results}
\begin{table*}[t]
\centering
\caption{\textbf{Comparison results on RVE-Bench.} The best and second best results are shown in \textbf{bold} and \underline{underlined} respectively. $^*$PC measures the consistency of unedited regions and is inapplicable to ICVG, whose output is a newly generated video segment rather than a modified version of the source.}
\label{tab:main_result}
\resizebox{\textwidth}{!}{%
\begin{tabular}{l|ccccc|cccccc}
\toprule
\textbf{Method}
  & \multicolumn{5}{c|}{\textbf{VLM Evaluation Score}}
  & \multicolumn{6}{c}{\textbf{Video Quality}} \\
\cmidrule(lr){2-6}\cmidrule(lr){7-12}
  & \textbf{EA$\uparrow$} & \textbf{PC$^*\uparrow$} & \textbf{GN$\uparrow$} & \textbf{GR$\uparrow$} & \textbf{Overall$\uparrow$}
  & \textbf{ViCLIP\textsubscript{T}$\uparrow$} & \textbf{SC$\uparrow$} & \textbf{BC$\uparrow$} & \textbf{OC$\uparrow$} & \textbf{TF$\uparrow$} & \textbf{MS$\uparrow$} \\
\midrule
\multicolumn{12}{l}{\textbf{Reasoning-Aware Video Editing}}  \\
\midrule
VACE~\cite{jiang2025vace}       & 2.6309 & \underline{6.7198} & \underline{6.6062} & \underline{7.3120} & 2.7195
           & 0.1658 & 0.955 & 0.954 & 0.151 & 0.973 & 0.985 \\
InsViE~\cite{wu2025insvie}     & 3.2111 & 6.4432 & 6.1679 & 6.2481 & 2.6281
           & 0.1688 & 0.952 & 0.952 & 0.156 & 0.941 & 0.974 \\
InsV2V~\cite{cheng2023consistent}     & 3.7815 & \textbf{7.9123} & 6.4580 & 7.2012 & \underline{3.5387}
           & 0.1731 & 0.956 & 0.963 & 0.167 & 0.966 & 0.980 \\
Omni-Video~\cite{tan2025omni} & \underline{4.8506} & 2.9716 & 6.5346 & 7.1519 & 3.1331
           & \underline{0.1747} & \underline{0.972} & \underline{0.971} & \underline{0.167} & \underline{0.979} & \underline{0.989} \\
\grayrow ReViSE & \textbf{6.1605} & 5.4086 & \textbf{6.6185} & \textbf{7.3123} & \textbf{4.6689}
           & \textbf{0.1771} & \textbf{0.981} & \textbf{0.978} & \textbf{0.174} & \textbf{0.983} & \textbf{0.991} \\
\midrule
\multicolumn{12}{l}{\textbf{In-Context V2V Generation}}  \\
\midrule
VACE~\cite{jiang2025vace}       & 3.3094 & -- & 5.9545 & 4.9703 & 2.9942
           & 0.2108 & 0.892 & 0.927 & 0.181 & 0.961 & 0.974 \\
InsViE~\cite{wu2025insvie}     & 3.0751 & -- & 4.9825 & 4.1136 & 2.2915
           & 0.2116 & 0.916 & 0.941 & 0.180 & 0.954 & 0.973 \\
InsV2V~\cite{cheng2023consistent}     & 3.6914 & -- & 5.8966 & 5.6431 & 3.3816
           & \underline{0.2153} & 0.847 & 0.921 & \textbf{0.196} & 0.962 & 0.974 \\
Omni-Video~\cite{tan2025omni} & \underline{5.1571} & -- & \underline{6.8123} & \textbf{7.4090} & \underline{3.8395}
           & 0.2148 & \underline{0.950} & \underline{0.957} & 0.187 & \underline{0.976} & \underline{0.986} \\
\grayrow ReViSE & \textbf{5.8235} & -- & \textbf{6.8476} & \underline{6.9645} & \textbf{3.9576}
           & \textbf{0.2154} & \textbf{0.952} & \textbf{0.961} & \underline{0.189} & \textbf{0.976} & \textbf{0.988} \\
\bottomrule
\end{tabular}}
\end{table*}
\subsubsection{Quantitative Results.}
As shown in Table~\ref{tab:main_result}, we evaluate five models on RVE-Bench across two complementary subsets: RAVE and ICVG.
On the RAVE subset, ReViSE achieves an Overall score of 4.6689, outperforming the Omni-Video by \textbf{49\%} (3.1331$\rightarrow$4.6689), with the most pronounced gains in EA (4.8506$\rightarrow$6.1605, \textbf{+27\%}) and PC (2.9716$\rightarrow$5.4086, \textbf{+82\%}). The ablation study (Table~\ref{tab:ablation_obj}) further shows that ReViSE surpasses the SFT-trained counterpart by \textbf{10\%} in Overall (4.2422$\rightarrow$4.6689), demonstrating that self-reflective learning substantially improves both instruction adherence and region preservation beyond supervised fine-tuning.
On the ICVG subset, ReViSE again outperforms Omni-Video in EA (5.1571$\rightarrow$5.8235, \textbf{+13\%}) and Overall score (3.8395$\rightarrow$3.9576). The slight GR decrease reflects the increased difficulty of maintaining visual realism when generating novel content without source-target correspondence constraints.
Compared with non-unified baselines, InsV2V and InsViE score higher in PC on RAVE, but their high PC largely reflects conservative edits that leave the source video mostly unchanged, which does not address the reasoning task. ReViSE achieves a balanced PC of 5.4086 alongside the highest EA, demonstrating that our approach improves reasoning-level editing without sacrificing preservation quality.

\noindent \textbf{Qualitative Results.}
\begin{figure*}[t]
  \centering
\includegraphics[width=0.95\linewidth]{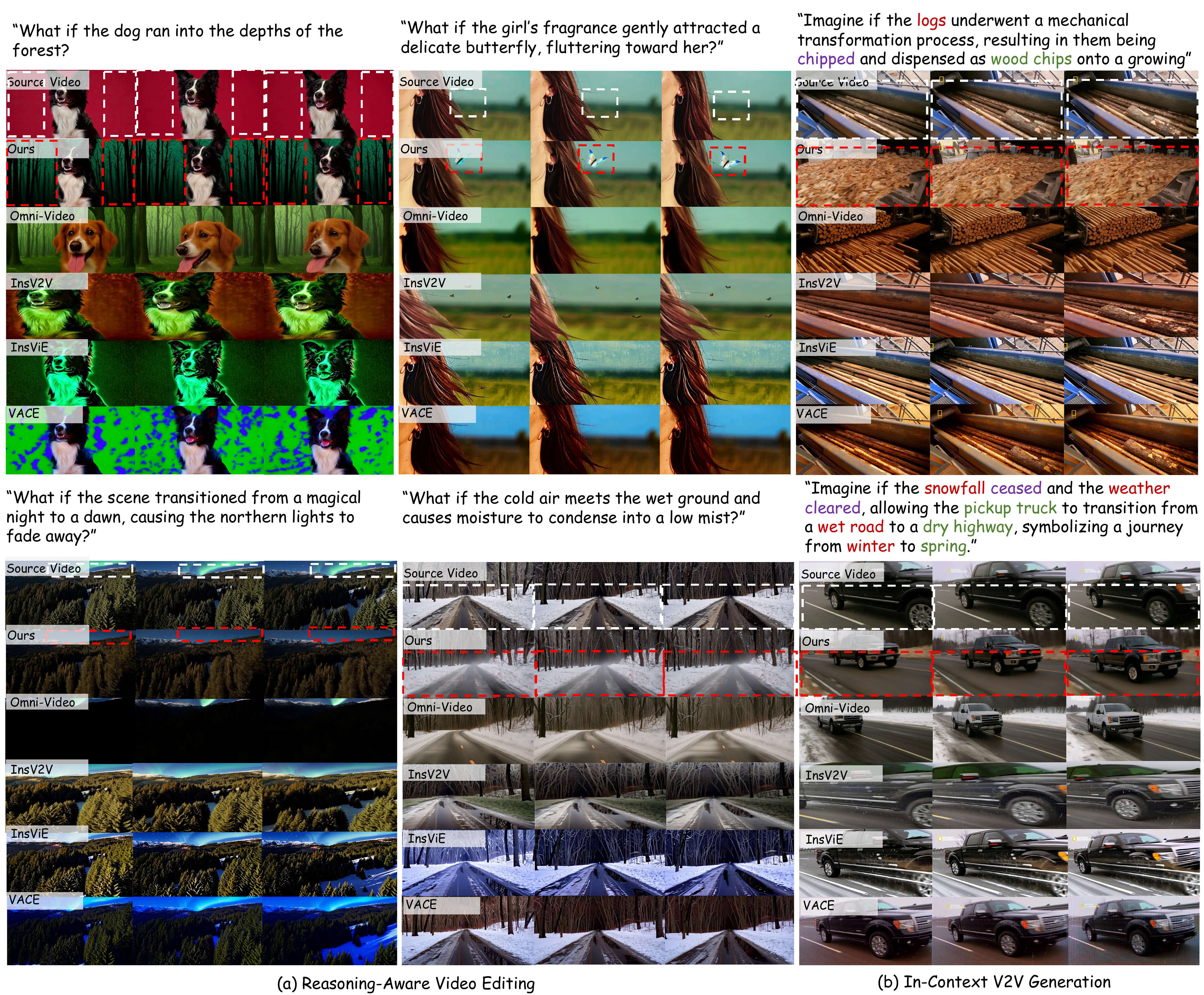}
   \caption{
   \textbf{Qualitative comparison of ReViSE and other baselines on RVE-Bench}, covering (a) Reasoning-Aware Video Editing (RAVE) and (b) In-Context V2V Generation (ICVG).
   White dashed boxes indicate regions where editing should occur; red boxes highlight regions where the editing is correctly performed.
   }
   \label{fig:quality}
\end{figure*}
Qualitative results in Figure~\ref{fig:quality} demonstrate ReViSE's effectiveness in
the RVE task. For instance, ReViSE successfully transforms a background into a ``depth
forest'' from the instruction ``ran into the depth forest,'' while maintaining visual
consistency and avoiding artifacts. When instructed to reflect the effects of cold air
meeting wet ground and causing moisture, ReViSE generates a foggy scene that logically
aligns with the physical process described in the instruction.
Figure~\ref{fig:quality}(b) showcases qualitative results for In-Context V2V
Generation, where instructions require rich contextual reasoning. For instance, the
instruction ``Imagine if the logs underwent a mechanical $\ldots$ wood chips onto a
growing pile'' demands understanding the transformation from logs to chips. ReViSE is
the only method that accurately interprets the instruction, generating a realistic video
of logs transforming into chips and accumulating into a growing pile, demonstrating its
unique ability to handle complex in-context generation tasks.

\noindent \textbf{User Study.}
\begin{figure*}[t]
  \centering
   \includegraphics[width=1.0\linewidth]{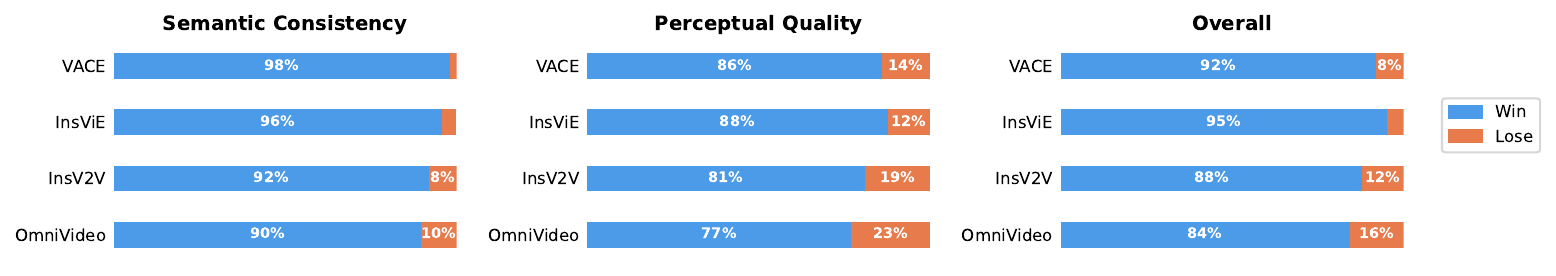}
   \caption{
   {User study results. We conduct 1-to-1 paired comparisons between ReViSE and each baseline; each rater labels the comparison as \textit{Win} or \textit{Lose}.}
   }
   \label{fig:user_study}
\end{figure*}
We randomly sample 50 RVE samples and conduct a human preference study with 15 experts. Each expert is shown a source video, a reasoning instruction, and the anonymized outputs of all five methods, and asked to rank them per criterion (\textit{Edit Accuracy}, \textit{Content Preservation}, \textit{Visual Quality}). Pairwise win rates are derived from the rankings. As shown in Figure~\ref{fig:user_study}, ReViSE achieves the highest win rate against all four baselines across all three criteria.

\noindent \textbf{Conventional Editing Results.}
To evaluate whether ReViSE retains conventional editing capability, we randomly selected $\sim$800 samples from the Ditto-1M dataset~\cite{bai2025scaling}. As shown in Table~\ref{tab:conventional_video_editing}, ReViSE outperforms all baselines in ViCLIP\textsubscript{T}, EA, GN, GR, and Overall, achieving a \textbf{36.7\%} gain in Overall (2.7871$\rightarrow$3.8109) over the previous best, demonstrating that ReViSE maintains strong conventional editing capability alongside its reasoning-informed editing.

\noindent \textbf{Cross-Dataset Results.}
We conduct additional experiments on VIE-Bench~\cite{mou2025instructx}, where standard instructions are rewritten by GPT-4o into reasoning-informed instructions. As shown in Table~\ref{tab:cross_dataset}, ReViSE achieves the best Overall (6.3819) and leads in EA (6.7312) and GN (7.3656), confirming generalization to unseen benchmarks under reasoning-informed settings.
\begin{table}[t]
\centering
\caption{Cross dataset comparison results on VIE-Bench~\cite{mou2025instructx}.}
\label{tab:cross_dataset}
\resizebox{\linewidth}{!}{%
\begin{tabular}{l|ccccc}
\toprule
\textbf{Method}& \textbf{EA$\uparrow$} & \textbf{PC$\uparrow$} & \textbf{GN$\uparrow$} &\textbf{GR$\uparrow$} & \textbf{Overall$\uparrow$}\\
\midrule
VACE~\cite{jiang2025vace} &6.5914  & \textbf{8.2366}&7.2151 & 8.1290&6.2858 \\
InsViE~\cite{wu2025insvie}&6.4624&8.2258  &7.2151  &8.1720 &6.1210\\
InsV2V~\cite{cheng2023consistent}& \underline{6.7065} & 8.1828 & 7.1398 & 8.1183 & \underline{6.3654}\\
Omni-Video~\cite{tan2025omni}&6.4731& 8.1290 & \underline{7.2473} &\underline{8.2473} & 6.2934\\
ReViSE (Ours) &\textbf{6.7312}& \underline{8.2258} & \textbf{7.3656} & \textbf{8.2473} & \textbf{6.3819}\\
\bottomrule
\end{tabular}}
\end{table}

\begin{table}[t]
\centering
\caption{Comparison results of conventional video editing on Ditto-1M~\cite{bai2025scaling}.}
\label{tab:conventional_video_editing}
\resizebox{\linewidth}{!}{%
\begin{tabular}{l|cccccc}
\toprule
\textbf{Method}& \textbf{ViCLIP\textsubscript{T}$\uparrow$} & \textbf{EA$\uparrow$} & \textbf{PC$\uparrow$} & \textbf{GN$\uparrow$} &\textbf{GR$\uparrow$} & \textbf{Overall$\uparrow$}\\
\midrule
VACE~\cite{jiang2025vace} &0.1721&1.4741  & \underline{5.7444}&5.9753 & 6.2383&1.3958 \\
InsViE~\cite{wu2025insvie}&0.1733&2.2185  &5.4185  &5.2420 &5.2840 &1.6960\\
InsV2V~\cite{cheng2023consistent} &0.1828& 3.2395 & \textbf{6.6247} & 5.7432 & 5.8864 & \underline{2.7871}\\
Omni-Video~\cite{tan2025omni} &\underline{0.1851}  &\underline{3.6901}& 3.1321 & \underline{6.5235} & \underline{7.0222} &2.5508\\
ReViSE (Ours) &\textbf{0.1877}& \textbf{5.0963} & 4.6852 & \textbf{6.5259} & \textbf{7.1543} & \textbf{3.8109}\\
\bottomrule
\end{tabular}}
\end{table}
\subsection{Ablation Study}
\noindent\textbf{Effect of Optimization Objective.}\quad
Table~\ref{tab:ablation_obj} compares training objective variants on the RAVE subset of RVE-Bench. While SFT and RWR~\cite{liu2025improving} both improve over the Omni-Video baseline, USO achieves the best Overall score. Specifically, SFT improves EA by only 1.8\%, whereas USO and RWR achieve gains of 27\% and 11\%, respectively. Although SFT obtains slightly higher GN and GR by focusing solely on reconstruction fidelity, it sacrifices EA and PC, the metrics most critical to reasoning-informed editing. RWR scales the flow-matching loss by a reward-derived weight $\alpha = P(\text{``Yes''})$, \ie, $\mathcal{L}_{\text{RWR}} = \alpha \cdot \mathcal{L}_{\text{FM}}$, but does not explicitly decouple edit generation from instruction satisfaction. USO instead jointly optimizes flow-matching and reasoning losses with independent gradients, achieving better editing performance.
\begin{table}[t]
\centering
\caption{Ablation results of training objectives of ReViSE on Reasoning-Informed Video Editing subset.}
\label{tab:ablation_obj}
\resizebox{\linewidth}{!}{%
\begin{tabular}{l|cccccc}
\toprule
\textbf{Method}& \textbf{ViCLIP\textsubscript{T}$\uparrow$} & \textbf{EA$\uparrow$} & \textbf{PC$\uparrow$} & \textbf{GN$\uparrow$} &\textbf{GR$\uparrow$} & \textbf{Overall$\uparrow$}\\
\midrule
Omni-Video~\cite{tan2025omni} &0.1747& 4.8506 & 2.9716 & 6.5346 & 7.1519 & 3.1331 \\
\midrule
\-w\ SFT &0.1721& \underline{5.9491} & 4.4334 & \textbf{6.9776}& \textbf{7.4969} & \underline{4.2422}\\
\-w\ RWR &\textbf{0.1781}& 5.3988 &\underline{4.5099} & 6.3370 & 6.9519 & 3.7554 \\
\-w\ USO (Ours) &\underline{0.1771} & \textbf{6.1605}&\textbf{5.4086} &\underline{6.6185} & \underline{7.3123} &\textbf{4.6689}\\
\bottomrule
\end{tabular}}
\end{table}

\begin{table}[t]
\centering
\caption{Performance comparison with different $\lambda$ settings of RVE-Bench.}
\label{tab:sigma}
\resizebox{\linewidth}{!}{%
\begin{tabular}{l|cccccc}
\toprule
\textbf{$\lambda$} & \textbf{ViCLIP\textsubscript{T}↑} & \textbf{EA↑} & \textbf{PC↑} & \textbf{GN↑} & \textbf{GR↑} & \textbf{Overall↑} \\
\midrule
0.10  & 0.1784 & 5.4247  & 4.5099 & 6.2407 & 6.7790 & 3.8174 \\
0.25 (Ours)  & \underline{0.1771} &\textbf{6.1605}&\textbf{5.4086} &\textbf{6.6185} & \textbf{7.3123} &\textbf{4.6689} \\
0.50  & \textbf{0.1801} & 5.6296 & 4.1790 & 6.5407 & 7.1617 & 3.9552 \\
0.75  & 0.1747 &6.0801  & 5.2303&  6.6133 & 7.1652 & 4.4114 \\
\bottomrule
\end{tabular}}
\end{table}

\noindent\textbf{Effect of Reasoning Loss Weight $\lambda$.}\quad
Table~\ref{tab:sigma} ablates the trade-off weight $\lambda$ between flow-matching and semantic losses.
$\lambda=0.25$ achieves the best Overall performance.
Too little semantic supervision at $\lambda=0.10$ depresses both EA and PC, indicating that the model fails to internalize instruction-level reasoning.
Interestingly, $\lambda=0.75$ partially recovers, with EA rising back to 6.0801 and Overall to 4.4114, yet ViCLIP\textsubscript{T} regresses to 0.1747, matching the baseline level, revealing that excessive semantic supervision ultimately compromises visual generation quality.
We therefore adopt $\lambda=0.25$ in all experiments.
\begin{table}[t]
\centering
\caption{Effect of $\sigma_\text{max}$ on reward signal quality and training coverage.}
\label{tab:sigma_ablation}
\resizebox{\linewidth}{!}{%
\begin{tabular}{lccccc}
\toprule
$\sigma_\text{max}$ & $0.1$ & $0.3$ & $0.5$ (Ours) & $0.7$ & $0.9$ \\
\midrule
Pearson $r$  & 0.476 & 0.458 & 0.451 & 0.389 & 0.201 \\
Trigger Rate & 3.5\% & 12.5\% & 25.0\% & 43.7\% & 75.0\% \\
\bottomrule
\end{tabular}}
\end{table}

\noindent\textbf{Effect of $\sigma_{\max}$.}\quad
Table~\ref{tab:sigma_ablation} examines how $\sigma_{\max}$ affects reward signal quality and training coverage.
Pearson $r$ decreases monotonically with $\sigma_{\max}$, confirming that heavier noise degrades the reliability of one-step denoising estimates.
Although $\sigma_{\max}=0.1$ yields the highest $r$, its trigger rate of only 3.5\% provides insufficient training coverage.
Raising $\sigma_{\max}$ to 0.5 reduces $r$ modestly to 0.451 while expanding coverage to 25.0\%, retaining 93.4\% of the clean-video upper bound ($r=0.483$) at a practical trigger rate.
Beyond $\sigma_{\max}=0.5$, reward quality deteriorates sharply.
We therefore select $\sigma_{\max}=0.5$ to balance signal fidelity and training coverage.
\section{Conclusion}
In this work, we introduced the Reason-Informed Video Editing (RVE) task, targeting the underexplored gap between literal instruction following and reasoning-grounded video editing.
To support this task, we constructed the RVE-Dataset and the RVE-Bench, providing large-scale training data and the first dedicated benchmark spanning diverse reasoning dimensions.
Building on this foundation, we proposed ReViSE, a self-reflective learning framework that uses the model's frozen internal VLM as an intrinsic differentiable reward, bridging the disconnect between VLM reasoning and video generation without external critics or labeled preference data.
Extensive experiments demonstrate that ReViSE achieves state-of-the-art performance across all reasoning categories on RVE-Bench and generalizes well to conventional video editing benchmarks.
{
    \small
    \bibliographystyle{ieeenat_fullname}
    \bibliography{main}
}
\clearpage
\appendix
\setcounter{page}{1}

\end{document}